\documentclass[conference]{IEEEtran}
\IEEEoverridecommandlockouts
\usepackage{cite}
\usepackage{amsmath,amssymb,amsfonts}
\usepackage{algorithmic}
\usepackage{graphicx}
\usepackage{textcomp}
\usepackage{xcolor}
\usepackage{multirow}
\def\BibTeX{{\rm B\kern-.05em{\sc i\kern-.025em b}\kern-.08em
    T\kern-.1667em\lower.7ex\hbox{E}\kern-.125emX}}
\begin{document}

\title{Global Intervention and Distillation for Federated Out-of-Distribution Generalization}

\author{
        \IEEEauthorblockN{ 
        Zhuang Qi\IEEEauthorrefmark{2}\textsuperscript{1}, 
        Runhui Zhang\IEEEauthorrefmark{2}\textsuperscript{1},  
        Lei Meng\IEEEauthorrefmark{1}\textsuperscript{2,1},
        Wei Wu\textsuperscript{1},
        Yachong Zhang\textsuperscript{1},
        and Xiangxu Meng\textsuperscript{1}
        }
    \IEEEauthorblockA{\textsuperscript{1} School of Software, Shandong University, Jinan, China\\}
    \IEEEauthorblockA{\textsuperscript{2} Shandong Research Institute of Industrial Technology, Jinan, China\\}

    Email: \{z\_qi, wu\_wei, zhangyachong\}@mail.sdu.edu.cn, zhangrunhuiz@gmail.com,  \{lmeng, mxx\}@sdu.edu.cn

\thanks{\IEEEauthorrefmark{2} indicates equal contributions.}
\thanks{\IEEEauthorrefmark{1} indicates corresponding author.}    
}

\maketitle

\begin{abstract}
Attribute skew in federated learning leads local models to focus on learning non-causal associations, guiding them towards inconsistent optimization directions, which inevitably results in performance degradation and unstable convergence. Existing methods typically leverage data augmentation to enhance sample diversity or employ knowledge distillation to learn invariant representations. However, the instability in the quality of generated data and the lack of domain information limit their performance on unseen samples. To address these issues, this paper presents a global intervention and distillation method, termed FedGID, which utilizes diverse attribute features for backdoor adjustment to break the spurious association between background and label. It includes two main modules, where the global intervention module adaptively decouples objects and backgrounds in images, injects background information into random samples to intervene in the sample distribution, which links backgrounds to all categories to prevent the model from treating background-label associations as causal. The global distillation module leverages a unified knowledge base to guide the representation learning of client models, preventing local models from overfitting to client-specific attributes. Experimental results on three datasets demonstrate that FedGID enhances the model's ability to focus on the main subjects in unseen data and outperforms existing methods in collaborative modeling.
\end{abstract}


\section{Introduction}
\label{sec:intro}


Federated out-of-distribution (OOD) generalization aims to leverage data with diverse attributes from multiple decentralized sources for collaborative modeling, ultimately obtaining a model capable of generalizing to unseen environments \cite{gao2022survey,liu2023cross,qi2023cross,li2020federated}. It enables parameter-level interaction between clients and the server without involving data sharing \cite{meng2024improving,qi2025cross,wang2024fednlr}. Despite its advantages in privacy protection, significant attribute skew across data sources often undermines the effectiveness of collaboration in federated learning \cite{liao2024foogd,qi2022clustering},. This is primarily because local models often overfit client-specific features, resulting in divergent optimization directions among models and leading to performance degradation in unseen scenarios.



To mitigate the out-of-distribution (OOD) problem, existing methods can be broadly classified into two categories: methods based on data augmentation and knowledge distillation. The former method primarily focuses on increasing the diversity of sample data to simulate various environments. These methods either train data generators in a federated learning manner or directly use pre-trained generative models to create data with different styles. For example, StableFDG~\cite{park2024stablefdg} and CCST~\cite{chen2023federated} share style information between clients to generate samples with diverse styles, which facilitates the model to learn causal relationships in complex scenarios. However, they often pose privacy risks and their performance is limited by the quality of generated data. The latter approach aims to guide the model to learn domain-invariant features, reducing the interference of irrelevant attributes on the model inference. Common approaches in this area of research include methods based on regularization and feature decoupling. For instance, DFL~\cite{luo2022disentangled} focuses on decoupling background and category features to enhance the attention to the main object of the image. However, relying on single-domain information for learning typically reduce the performance in extracting invariant representations in unseen domains.

\begin{figure}[t]
\centering
\includegraphics[width=0.9\linewidth]{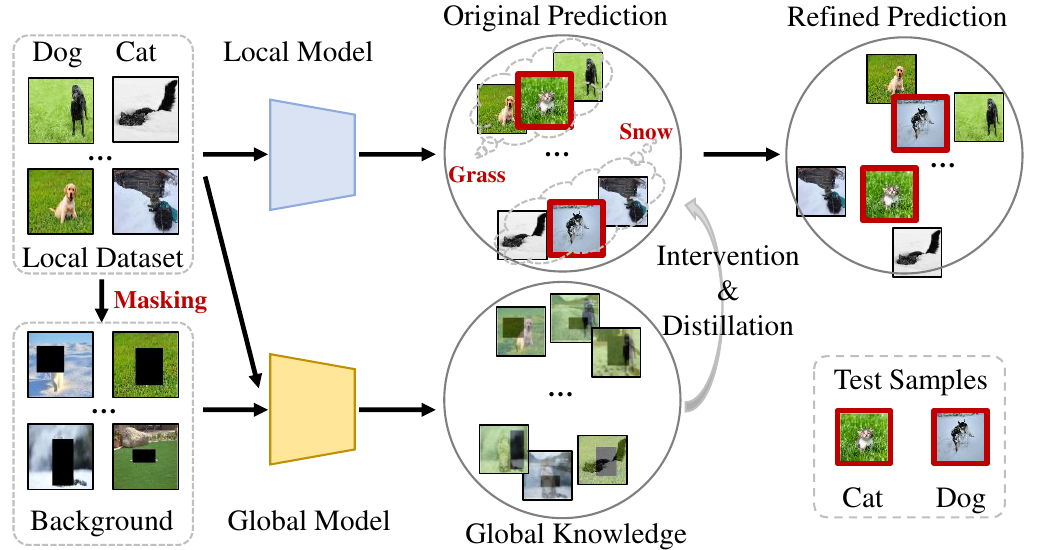}
\vspace{-0.3cm}
\caption{Motivation of the proposed FedGID. It employs background masking to remove associations between target categories and background attributes without sharing information across clients. Additionally, it uses global knowledge distillation to align feature distributions among clients, enhancing the model's generalization on biased data.
} 
\vspace{-0.5cm}
\label{fig1}
\end{figure}

To address these issues, this paper proposes a federated out-of-distribution generalization method that integrates global intervention and knowledge distillation, termed FedGID, as illustrated in Figure \ref{fig1}. Compared to traditional approaches, FedGID leverages inter-class background information to intervene in sample distributions and employs backdoor adjustment to effectively eliminate spurious associations between background factors and labels. FedGID comprises two main modules: the global intervention (GI) module and the global distillation (GD) module. Specifically, the GI module dynamically decouples target objects from background information in images, injecting background features from different categories into the original samples. This establishes associations between backgrounds and all categories, preventing the model from mistakenly treating background as a causal factor. Meanwhile, the GD module aligns the local features of client models with the generalized features of the global model, utilizing the guidance of a unified knowledge base to ensure that client models learn consistent representations across clients, which  avoids overfitting to client-specific attributes. This approach not only enhances the model's robustness to background variations but also serves as a plug-and-play module that can be easily integrated into existing methods.



Extensive experiments were conducted on three datasets in terms of performance comparisons, ablation studies of key modules, case study of visual attention on key regions, and error analysis. The results show that FedGID enhances the model's attention to objects in unseen data and improves the effectiveness of collaborative learning. In summary, this paper makes two main contributions:
\begin{itemize}
    \item This paper presents a model-agnostic global intervention and distillation mechanism, termed FedGID. To the best of our knowledge, it is the first method that leverages background factors to intervene in the data distribution in federated learning, which can break the spurious association between background and labels.

    \item The proposed global intervention module is a plug-and-play component that can be seamlessly integrated into various methods without altering their main architecture, which facilitates the enhancement of their capacity to discern causal relationships.
    
\end{itemize}

\section{Related Work}
\subsection{Data Augmentation-based Methods}

%
Attribute skew across clients typically leads the model to overfit domain-specific attributes, which reinforces spurious associations between these attributes and labels. To address this issue, data augmentation-based methods~\cite{shenaj2023learning,liu2021feddg,xu2023federated} aim to enhance the model's generalization to unseen attributes by increasing the diversity of data attributes. These methods either train data generators to produce new samples or directly use pre-trained diffusion models to enhance sample diversity. The former method typically exchanges local information between clients to generate new samples from different domains. For example, FIST~\cite{nguyen2024fisc} and StableFDG~\cite{park2024stablefdg} generate a series of data with different styles by sharing style information between clients. The latter methods generate samples that meet specific requirements based on prompt information. However, these methods may raise privacy concerns, and the quality gap between the generated data and the original samples can limit the model's performance.

\subsection{Knowledge Distillation-based Methods}

%

Knowledge distillation-based methods aim to leverage generalized knowledge to guide local models in learning shared features that are independent of attributes, which improves the model's generalization to unseen data. They either regularize to align representations from different sources, learning consistent representations across various contexts~\cite{qi2024attentive,wang2023dafkd,qi20241cross}, or decouple invariant features in the latent space to reduce the interference of confounding factors~\cite{yu2023contrastive}. For example, FedProc~\cite{mu2023fedproc} and FPL~\cite{fpl} use prototypes to regularize the representation learning of all clients, guiding them towards the same space. MCGDM~\cite{MCGDM} avoids overfitting within a single domain by using both intra-domain gradient matching and cross-domain gradient matching. FedIIR~\cite{fediir} implicitly learns invariant relationships by leveraging prediction inconsistency and gradient alignment across clients. Despite these methods yield performance improvements, the reliance on single-domain information restricts the model's transferability across out-of-distribution domains.

\begin{figure}[t]
\centering
\includegraphics[width=0.8\linewidth]{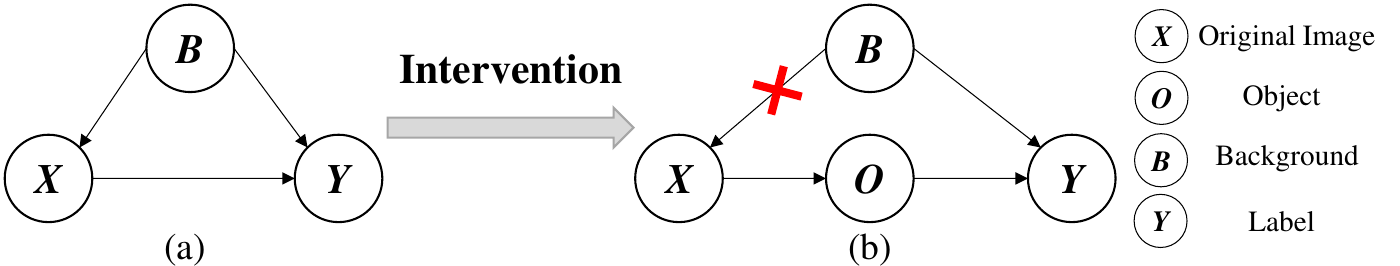}
\vspace{-0.3cm}
\caption{Illustration of the structural causal graph with (b) and without (a) intervention. It eliminates the association between the background (B) and the label (Y), enabling the model to establish the connection between the image (X) and the label (Y) by focusing on the main object (O).} 
\label{fig6}
\vspace{-0.3cm}
\end{figure}

\section{Method}

\subsection{Overall Framework}

\begin{figure*}[t]
\centering

\includegraphics[width=0.9\linewidth]{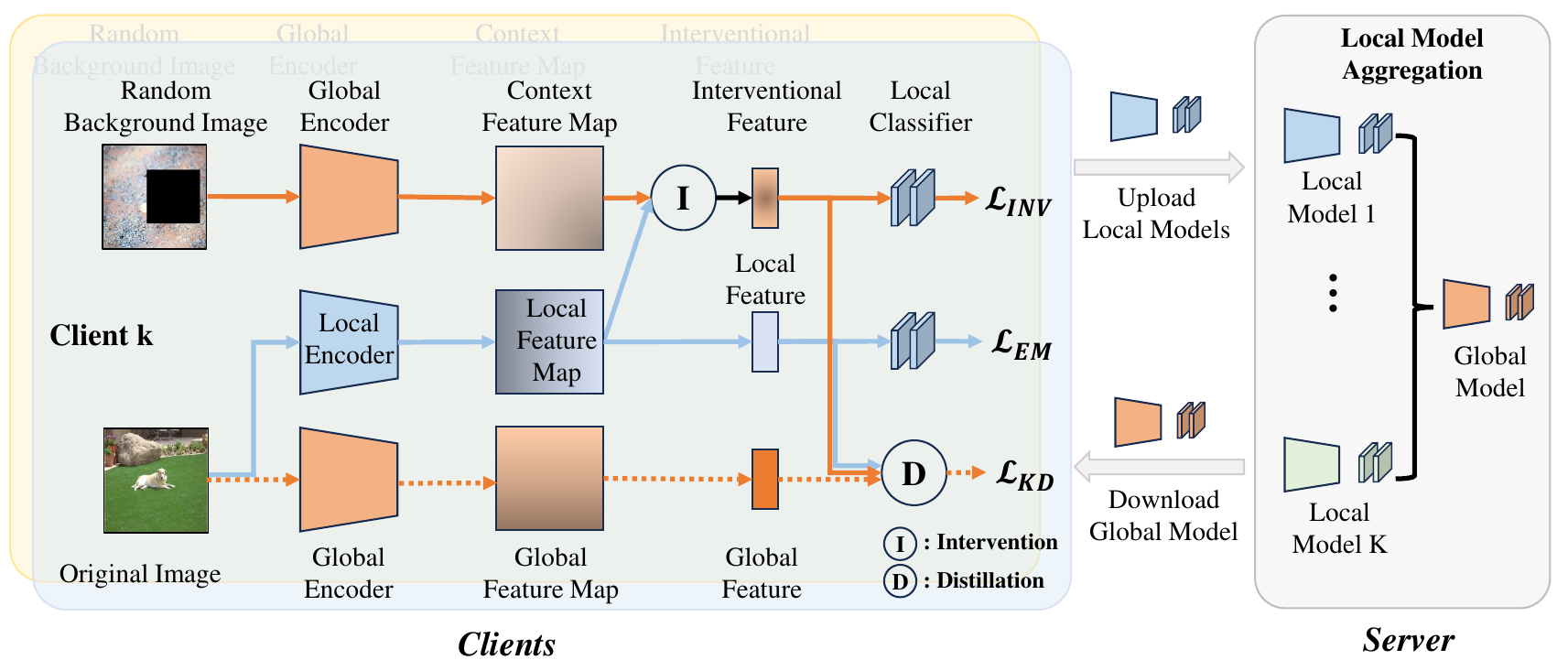}
\vspace{-0.3cm}
\caption{Illustration of the framework of FedGID. It consists of two main modules, including the global intervention module and the global distillation module. The former performs backdoor adjustment to intervene in the attribute distribution by fusing background information. The latter employs the global knowledge to build unified feature space across clients.} 
\label{fig2}
\vspace{-0.3cm}
\end{figure*}

This work proposes a global intervention and distillation method in federated learning, called FedGID, which enhances the model's generalization ability to unseen scenarios. Specifically, FedGID consists of two main modules, as shown in Figure \ref{fig2}. The global intervention module aims to introduce diversified information to fuse features from different attributes, which strengthens the model's adaptability to various changes. The global distillation module guides local models to learn shared features while disregarding attribute-specific features.

\subsection{Global Intervention (GI) Module}




Out-of-distribution (OOD) scenarios often lead models to form spurious correlations ($B \rightarrow X \rightarrow Y$) between background factors ($B$) and labels ($Y$) \cite{eccv,liu2022prompt,wang2024causal}, limiting their ability to generalize to unseen data. Therefore, the GI module addresses this by fusing inter-class background information to focus on causal relationships ($X\rightarrow O \rightarrow Y$, where $O$ represents object), as shown in Figure \ref{fig6}.

To achieve this, the GI module consists of two stages: the background extraction stage and the background-driven intervention stage. In the background extraction stage, the input \(X\) is decomposed into object \(O\) and background \(B\) (\(X = O + B\)). Specifically, a pre-trained DINO model~\cite{liu2023grounding} is used to detect and mask the object \(O\), isolating the background \(B\). This process can be expressed as:

\begin{equation}
  \theta= \begin{bmatrix}
  x_1 & x_2 \\
  y_1 & y_2
\end{bmatrix} = DINO(X, T),
\end{equation}
\begin{equation}
I_{B}=I \odot 1_{(x, y) \notin\left[x_{1}, x_{2}\right] \times\left[y_{1}, y_{2}\right]},
\end{equation}
where $I_B$ is the background image, and $T$ is the label name of the image $X$. $\theta$ is the coordinate matrix, with $(x_1, y_1)$ as the top-left and $(x_2, y_2)$ as the bottom-right corner. $\odot$ denotes the Hadamard product. $1_{(x, y) \notin [x_1, x_2] \times [y_1, y_2]}$ is an indicator function, which is 1 outside the bounding box and 0 inside.


Moreover, the background-driven intervention stage aims to 
inject the random background feature $f_{B}^{random}$ into the original image feature $f_{I}$, ensuring that each background is associated with multiple classes. This can be expressed as: 
\begin{equation}
    f_{INV}= \alpha * f_{I} \oplus (1-\alpha) * f_{B}^{random},
\end{equation}
\begin{equation}
    f_{I}=E_{L}(I), f_{B}^{random}=E_G(I_{B}^{random}),
\end{equation}
where $E_{L}(\cdot)$ and $E_G(\cdot)$ are local and global feature encoders. To simplify the description, client id are omitted. $\alpha$ is a hyperparameter. $f_{INV}$ denotes the interventional feature that share the same label as $f_{I}$. To effectively intervene in model training, we design the intervention classification loss $L_{GI}$,
\begin{equation}
    \mathcal{L}_{\mathrm{GI}}=-\frac{1}{N} \sum_{i=1}^{N} \sum_{c=1}^{C} y_{c}^{(i)} \log \left(\frac{e^{z_{c}^{(i)}}}{\sum_{j=1}^{C} e^{z_{j}^{(i)}}}\right)
\end{equation}
where $ N $ is the batch size, $ C $ is the number of classes, $ z_c=F_L(f_{INV}) $, $F_L(\cdot)$ is the client classifier, $ y_c^{(i)} $ and $ z_c^{(i)} $ represent the one-hot encoded label and predicted score for class $ c $ of the feature $f_{INV}$, respectively.



   

\subsection{Global Distillation (GD) Module}


To build a unified knowledge base across models and prevent the interference of client-specific attributes, the Global Distillation (GD) module uses global knowledge as a soft regularizer to guide local representation learning, mitigating overfitting to non-causal factors and promoting collaboration between heterologous models. Specifically, it uses Kullback-Leibler (KL) divergence to encourage the local model to adjust its feature representations $f_I$ based on the global features $f_G$. The process can be expressed as: 
\begin{equation}
     \mathcal{L}_{I \parallel G} = \text{KL}(f_I \parallel f_G)
\end{equation}
where $f_G=E_G(I)$. Meanwhile, to enhance the generalization capability of the model across different environments, the GD module also mitigates the gap between the interventional features $f_{INV}$ and global features, i.e., 
\begin{equation}
     \mathcal{L}_{INV \parallel G} = \text{KL}(f_{INV} \parallel f_G)
\end{equation}
Therefore, the GD module indirectly mitigates the gap in heterologous feature spaces by reducing the difference between global and local features. Its optimization objective can be expressed as:
\begin{equation}
    \mathcal{L}_{GD} = \mathcal{L}_{I \parallel G} + \mathcal{L}_{INV \parallel G} 
\end{equation}
Additionally, to ensure the model's classification performance on the samples, this study employs the standard cross-entropy loss to optimize the local model:
 \begin{equation}
  \mathcal{L}_{EM} = CE(F_a(f_I), y)
 \end{equation}
where $F_a(\cdot)$ represents the local classifier, and $y$ denotes the label of the original image $I$.

%


\subsection{Training Strategy of FedGID}

FedGID aims to enhance the model's attention to image objects and improve the performance of collaborative modeling by global intervention and distillation. Its overall optimization objective can be express as:
\begin{equation}
    \mathcal{L}_{total} = E_{(x,y)\sim D_{local}}[\mathcal{L}_{EM} + \mathcal{L}_{GI} + \lambda * \mathcal{L}_{GD}] 
\end{equation}
where $\lambda$ is a weighted parameter.


\begin{table*}[t]
    \centering
     \caption{Performance comparison between FedGID with baselines on  COLORMNIST,NICO-Animal and NICO-Vehicle. All methods were executed across three trials, with both the mean and standard deviation being reported.}
     \resizebox{0.85\textwidth}{!}{
     \setlength{\tabcolsep}{3mm}{
    \begin{tabular}{c|c|c|c|c|c|c}
    \hline   
      \multicolumn{1}{c|}{\multirow{2}*{\textbf{Methods}}}& \multicolumn{2}{c|}{\textbf{NICO-Animal}} &
      \multicolumn{2}{c|}
      {\textbf{NICO-Vehicle}}&\multicolumn{2}{c}{\textbf{ColorMNIST}}  \\
     \cline{2-7}
     \multicolumn{1}{c|}{}&A7&B7&A7&B7&\(\beta\)=0.1&\(\beta\)=0.5\\
     \hline
     FedAvg (AISTATS'17)~\cite{fedavg}&44.38±0.6&52.75±0.6 &65.28±0.4&59.05±0.2&89.39±0.6&89.97±0.5\\
     Fedprox (MLSys'20)~\cite{fedprox}&44.55±0.9&51.99±0.9&65.36±0.6&57.50±0.8&86.84±0.4&87.77±0.3\\
     MOON (CVPR'21)~\cite{moon}&45.53±0.4&53.66±0.9&65.94±0.5&59.63±0.5&91.68±0.8&91.15±0.2\\
     FPL (CVPR'23)~\cite{fpl}&47.76±0.5&55.39±0.2&68.51±0.7&61.76±0.6&92.94±0.3&95.79±0.8\\
     FedIIR (ICML'23)~\cite{fediir}&46.40±0.9&52.82±0.7&63.64±0.9&56.18±0.4&89.69±0.8&90.23±0.9\\
      FedHeal (CVPR‘24)~\cite{Fedheal}&42.32±1.0&52.80±0.6&64.00±0.5&56.25±0.7&92.02±0.1&87.66±0.6\\
     MCGDM (AAAI'24)~\cite{MCGDM}&47.96±0.8&54.53±0.5&66.84±0.4&59.59±0.9&89.45±0.5&95.40±0.9\\
     
     \hline
     FedGID$_\mathrm{FedAvg}$&48.54±0.2&55.32±0.5&67.67±0.2&61.79±0.4&92.83±0.4&93.98±0.5\\
     FedGID$_\mathrm{MOON}$&47.84±0.4&56.74±0.3&68.57±0.4&61.56±0.7&94.09±0.3&94.43±0.7\\
     FedGID$_\mathrm{FPL}$&\textbf{49.11±0.3}&\textbf{56.84±0.9}&\textbf{69.39±0.9}&\textbf{62.78±0.3}&\textbf{94.88±0.5}&\textbf{96.21±0.6}\\
     \hline
     \end{tabular}}}

    \label{tab:comparison}
\end{table*}

\begin{table}[h]
    \centering
    \vspace{-0.2cm}
    \caption{Statistics of COLORMNIST,NICO-Animal and NICO-Vehicle datasets used in experiments,where A7 represents to the data from the first seven backgrounds of each class used as the training set, while B7 represents to the data from the last seven backgrounds of each class used as the training set.}
    \label{tab1}
    \resizebox{0.4\textwidth}{!}{
    \begin{tabular}{c|c|c|c } 
    \hline
     \textbf{Datasets} & \textbf{\#Class} & \textbf{\#Training} & \textbf{\#Testing} \\ 
    \hline
    NICO-Animal (A7) & 10 & 10633 & 2443 \\
    \hline
    NICO-Animal (B7) & 10 & 8311 & 4765 \\
    \hline
    NICO-Vehicle (A7) & 10 & 8027 & 3626 \\
    \hline
    NICO-Vehicle (B7) & 10 & 8352 & 3301 \\
    \hline
    COLORMNIST & 10 & 60000 & 10000 \\
    \hline
  \end{tabular}}
  \vspace{-0.1cm}
\end{table}

\section{EXPERIMENTS}
\subsection{Experiment Settings}

\subsubsection{Datasets}

Following existing work, experiments are conducted on three commonly used datasets, NICO-Animal \cite{NICO_Causal}, NICO-Vehicle~\cite{NICO_Causal} and ColorMNIST~\cite{mnist}. Detailed statistical information about these datasets is provided in Table \ref{tab1}.

\subsubsection{Network Architecture}

To make a fair comparison, all methods use the same network architecture. In FedGID, a consistent architecture is employed to recognize both raw images and images processed with Grounding DINO. Based on previous studies~\cite{moon,liu2021feddg}, ResNet-18~\cite{he2016deep} is chosen as the backbone for NICO-Animal and NICO-Vehicle datasets, and a SimpleCNN with one convolutional layer and two fully connected layers is applied to the ColorMNIST dataset.

\subsubsection{Hyper-parameter Settings}

In the experiments, local training epochs per global round were set to 10 for the NICO-Animal and NICO-Vehicle datasets, and 5 for ColorMNIST. The number of communication rounds was 50, with 7 clients for NICO-Animal and NICO-Vehicle, and 5 clients for ColorMNIST. The client sampling fraction was 1.0, and SGD was used as the optimizer. During local training, weight decay was 0.01, batch size was 64, and initial learning rates were 0.01 for NICO-Animal and NICO-Vehicle, and 0.005 for ColorMNIST, the parameter $\lambda$ is selected from $\{0.1,1,5,10\}$. The Dirichlet parameter $\beta$ is set to 0.1 and 0.5 for ColorMNIST. For other methods, hyperparameters followed those outlined in the respective papers.


\subsection{Performance Comparison}
 This section compared the FedGID method with seven state-of-the-art (SOTA) methods, including FedAvg \cite{fedavg}, FedProx \cite{fedprox}, MOON \cite{moon}, FPL \cite{fpl}, FedIIR \cite{fediir}, FedHeal \cite{Fedheal}, and MCGDM \cite{MCGDM}. Furthermore, we incorporate the GI module into FPL and MOON, modeling their knowledge distillation modules as instances of the GD module, resulting in FedGID$_{\mathrm{MOON}}$ and FedGID$_{\mathrm{FPL}}$. The following results can be derived from Table \ref{tab:comparison}.
\begin{itemize}[]
\item \textbf{FedGID$_{\mathrm{FedAvg}}$, FedGID$_{\mathrm{MOON}}$ and FedGID$_{\mathrm{FPL}}$ achieve significant improvements across all cases} compared to baselines, highlighting the model-agnostic nature of the FedGID method and the effectiveness of the global intervention and distillation mechanism.

\item  \textbf{FedGID consistently outperforms other methods in image classification}, which is understandable, as it alleviates the spurious correlation between background factors and labels, allowing the model to better learn causal representations.
\item  \textbf{FedGID uses causal intervention to break the spurious correlation between background and labels within clients and combines knowledge distillation to create a consistent knowledge base across clients.} It addresses the federated OOD problem from both intra-client and inter-client perspectives, achieving better results than single-perspective methods, such as FedIIR, FPL.
\end{itemize}

\begin{table}[h]
    \centering
    \vspace{-0.5cm}
     \caption{Ablation study on the effectiveness of different components of FedGID on the NICO-Animal and NICO-Vehicle.}
     \resizebox{0.45\textwidth}{!}{
    \begin{tabular}{c|c|c|c|c}
         \hline
         \multicolumn{1}{c|}{\multirow{2}*{}} & \multicolumn{2}{c|}{\textbf{NICO-Animal}} & \multicolumn{2}{c}{\textbf{NICO-Vehicle}} \\
         \cline{2-5}
         \multicolumn{1}{c|}{}&A7&B7&A7&B7\\
        \hline
         \textbf{FedAvg}&44.38±0.6&52.75±0.6&65.28±0.4&59.05±0.2\\
        \hline
        \textbf{ + $\mathrm{GD}$ ${\mathrm{}}$}&46.04±0.5&53.99±0.3&66.21±0.7&59.58±0.6\\
        \textbf{+ $\mathrm{GI}_{\mathrm{F}}$ ${}$}&46.87±0.7&54.06±0.3&66.65±0.9&59.67±0.6\\
       \textbf{+ $\mathrm{GI}_{\mathrm{FM}}$ ${}$}&47.02±0.7&54.19±0.9&67.32±0.9&59.88±0.7\\
        \textbf{ + $\mathrm{GI}_{\mathrm{F}}$ + $\mathrm{GD}_{\mathrm{}}$}&48.13±0.6&55.20±0.7&66.87±0.3&61.45±0.7\\
        \textbf{+ $\mathrm{GI}_{\mathrm{FM}}$ + $\mathrm{GD}_{\mathrm{}}$}&\textbf{48.54±0.4}&\textbf{55.32±0.5}&\textbf{67.67±0.2}&\textbf{61.79±0.4}\\
        
         \hline
    \end{tabular}}
    \label{tab:ablation}
    \vspace{-0.3cm}
\end{table}

\begin{figure*}[t]
\centering
\includegraphics[width=0.95\linewidth]{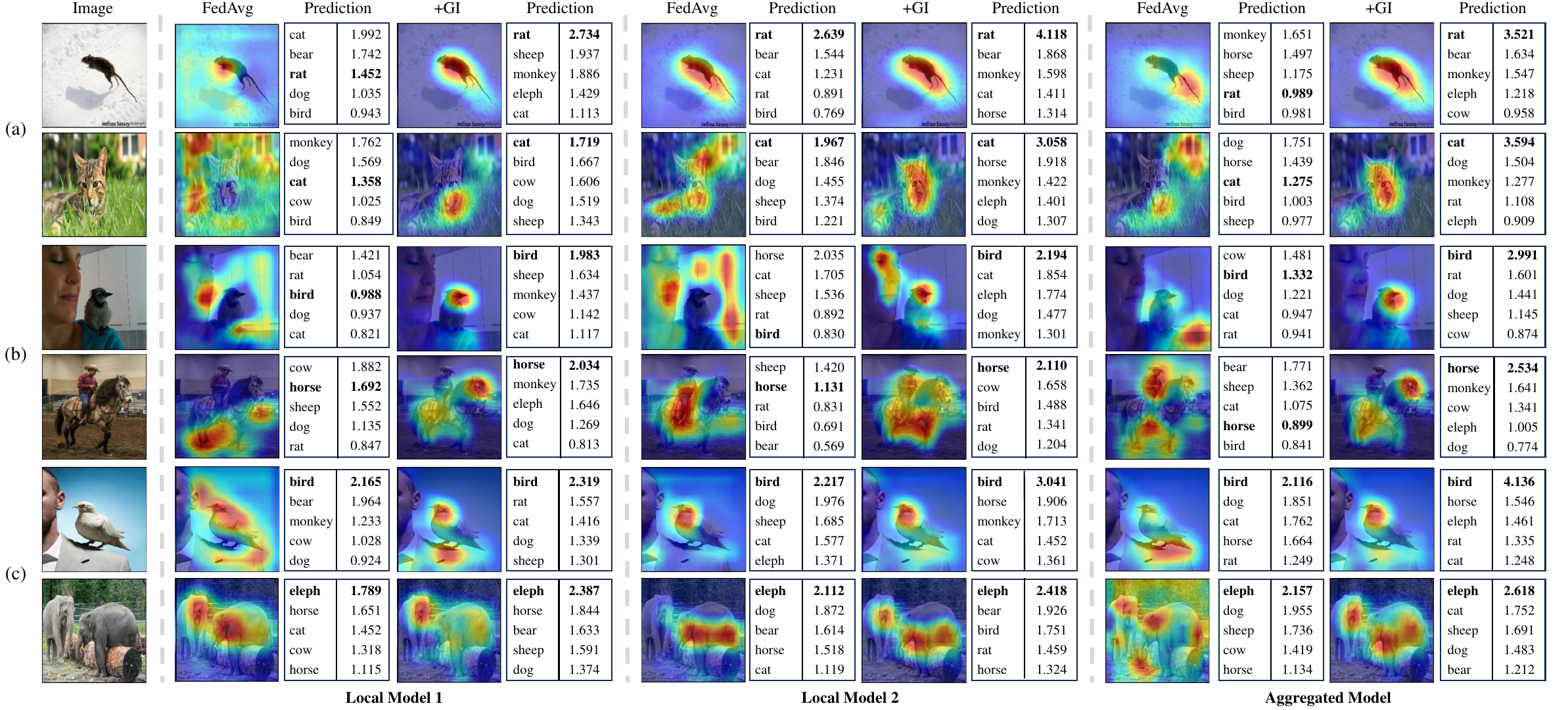}
\vspace{-0.3cm}
\caption{Visualization of the visual Attention. (a) The GI module corrects errors in individual clients. (b) The GI module improves the aggregated model by correcting errors in each client, even when both clients make mistakes. (c) The GI module increases the model's confidence in the ground-truth.} 
\label{fig3}
\vspace{-0.3cm}
\end{figure*}
\subsection{Ablation Study}
 This section further investigates the effectiveness of different modules within the FedGID framework. The results are summarized in Table \ref{tab:ablation}.
\begin{itemize}[]
\item  \textbf{Incorporating the global distillation module enhances the alignment between local and global features}, facilitating the transfer of knowledge across heterologous models and improving performance on diverse datasets.

\item \textbf{The GI module uses causal intervention to address spurious correlations between background and labels}, improving the model’s ability to generalize across clients and reducing overfitting to client-specific attributes.

\item \textbf{Intervention at the feature map level $GI_{FM}$ typically yields better performance than at the feature level $GI_{F}$.} This is because feature map-level intervention can retain more raw information while enhancing the model's understanding of the relationships between different backgrounds and objects.

\end{itemize}

\subsection{Case Study}

\subsubsection{Visual Attention Visualization}
This section analyzes the effectiveness of the global intervention mechanisms. Figure \ref{fig3} shows the model outputs of FedAvg and GI$_{FedAvg}$, along with visual attention based on GradCAM \cite{gradcam}. It is clear that \textbf{the GI module enhances the model's ability to focus on causal regions.} Specifically, \textbf{the GI module reduces irrelevant background noise to help the model correct prediction errors from the FedAvg.} As shown in Figure \ref{fig3}(a), erroneous predictions from individual clients can interfere with collaborative modeling, the GI module corrects these errors, which can improve the performance of the aggregated model. In Figure \ref{fig3}(b), despite both client models of FedAvg make incorrect predictions for all images, the GI module is able to correct all of these errors.
Figure \ref{fig3}(c) shows that the GI module further improves the prediction confidence of both client and global models, thereby increasing the prediction gap between ground-truth and other classes.


\subsubsection{Effectiveness analysis of heterogeneous feature alignment} 
\begin{figure}[t]
\centering
\includegraphics[width=0.9\linewidth]{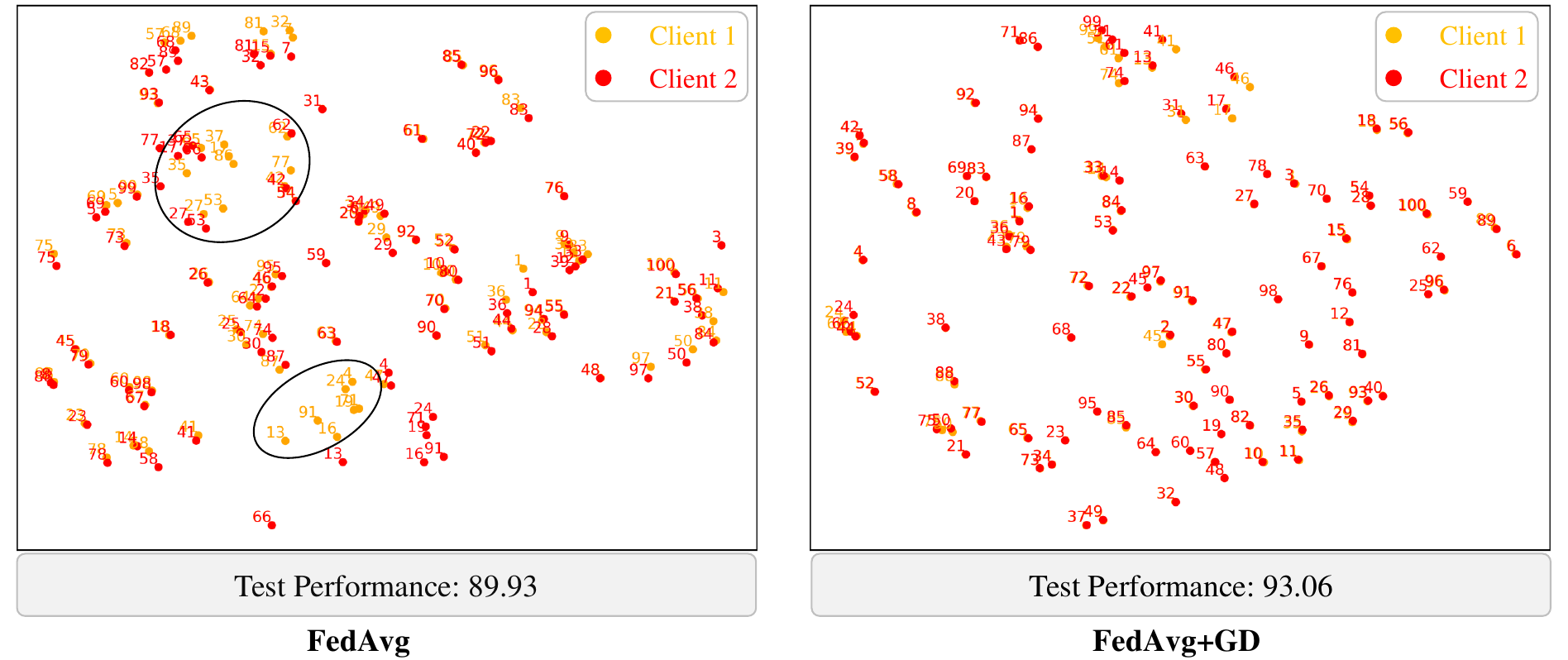}
\vspace{-0.2cm}
\caption{Visualization of the heterogeneous feature distribution for FedAvg and its version with the GD module. The GD module can mitigate the discrepancies between features from different sources.} 
\label{fig4}
\vspace{-0.4cm}
\end{figure}

This section aims to explore the effectiveness of the global distillation module for heterogeneous feature alignment on the ColorMNIST. 
Figure \ref{fig4} shows the effect of heterogeneous feature alignment before and after the use of the GD module. We randomly selected 100 test samples, with yellow and red points representing the predictions of Client 1 and Client 2 models, respectively. Visually, \textbf{the GD module significantly reduces the distribution difference in the heterogeneous feature space.} The two originally dispersed data distributions become much closer after applying the GD module, indicating a significant improvement in the collaboration between the two models. This feature alignment optimization helps enhance the overall collaboration performance of multi-source models.

\section{Conclusion}

To mitigate ill-posed aggregation arising from attribute skew, this paper proposes a global intervention and distillation method, termed FedGID. It utilizes backdoor adjustment to break the spurious associations between background factors and labels. Moreover, it employs global knowledge distillation to regulate local representation learning, effectively preventing overfitting to specific attribute features and bridging representation inconsistencies across clients. Experimental results show that global intervention and distillation improve the performance in focusing on the main subjects in unseen data and enhance collaboration across multiple clients.

Despite the impressive performance of FedGID, two directions remain worth exploring. Firstly, applying stronger representation learning can help in cases where attribute features can't be separated \cite{chen2023class,SeeDRec,meng2019learning,qi2021iterative,wang2022causal,chen2023classs,lin2020multi}. Secondly, extending FedGID to more challenging scenarios is a promising direction \cite{ma2023cross,li2024cross,wang2023multi,PDRec,qi2024machine,meng2020heterogeneous,TriCDR,li2022unsupervised,wang2024modeling,meng2017towards,li2021comparative}.

\section{Acknowledgments}
This work is supported in part by the TaiShan Scholars Program (Grant no. tsqn202211289), the Shandong Province Excellent Young Scientists Fund Program (Overseas) (Grant no. 2022HWYQ-048).

\bibliographystyle{IEEEbib}
\bibliography{icme2025references}


\end{document}